%% file: arxiv.tex
\documentclass[letterpaper]{article} 
\usepackage{for_arxiv}  
\usepackage{times}  
\usepackage{helvet}  
\usepackage{courier}  
\usepackage[hyphens]{url}  
\usepackage{graphicx} 
\usepackage{natbib}
\usepackage{caption}
\usepackage{algorithm}
\usepackage{algorithmicx}
\usepackage[noend]{algpseudocode}
\usepackage{newfloat}
\usepackage{listings}
\DeclareCaptionStyle{ruled}{labelfont=normalfont,labelsep=colon,strut=off} 
\floatstyle{ruled}
\newfloat{listing}{tb}{lst}{}
\floatname{listing}{Listing}
 
\usepackage{amsfonts,amsmath}
\usepackage{import}
\usepackage{tikz,pgffor}
\usepackage{ifthen}
\usepackage{booktabs}
\usepackage{hyperref}
\usepackage{balance}
\hypersetup{
	colorlinks=true,
	linkcolor=black,
	citecolor=blue}
\usetikzlibrary{arrows,backgrounds,calc}
\usetikzlibrary{automata,positioning,arrows,shapes,math,arrows.meta,decorations.pathmorphing}
\tikzstyle{min}=[thick,circle,draw,minimum size=1.4em,inner sep=0em,text centered]
\tikzstyle{dec}=[circle,draw,fill,minimum size=.8ex,inner sep=0em]
\usepackage{scalerel}

\newtheorem{theorem}{Theorem}

\usepackage[noend]{algpseudocode}
\algrenewcommand\algorithmiccomment[1]{\hfill\textcolor{blue}{\commentsymbol{} #1}}

\input{macros}

\title{Multiple Mean-Payoff Optimization under Local Stability Constraints}

\author{
David Kla\v{s}ka,
Anton\'{\i}n Ku\v{c}era, 
Vojt{\v{e}}ch K\r{u}r,
V\'{\i}t Musil,  
Vojt\v{e}ch \v{R}eh\'{a}k}

\affiliations{
Masaryk University, Brno, Czechia\\
\{david.klaska\}@mail.muni.cz,
\{tony, musil, rehak\}@fi.muni.cz
}

\begin{document}

\maketitle

\import{sections}{abstract}
    

\import{sections}{intro}
\import{sections}{model}
\import{sections}{problem}
\import{sections}{algorithm}
\import{sections}{experiments}
\import{sections}{conclusion}

\input{arxiv.bbl}
\balance

\clearpage
\appendix

\twocolumn[
\vbox{%
    \hsize\textwidth%
    \linewidth\hsize%
    \vskip 0.625in minus 0.125in%
    \centering%
    {\LARGE\bf Supplementary Material \par}%
    \vskip 2em plus 2fil%
}]

\import{sections}{appendix}

\end{document}

%% file: macros.tex
\usepackage{tikz,pgffor}
\usetikzlibrary{arrows,backgrounds,calc}
\usetikzlibrary{automata,positioning,arrows,shapes,math,arrows.meta,decorations.pathmorphing}
\tikzstyle{min}=[thick,circle,draw,minimum size=1.3em,inner sep=0em,text centered]

\newcommand{\Nset}{\mathbb{N}}

\newcommand{\Qset}{\mathbb{Q}}

\newcommand{\Rset}{\mathbb{R}}
\newcommand{\Exp}{\mathbb{E}}

\newcommand{\Inv}{\mathbb{I}}

\newcommand{\GMP}{\mathrm{GMP}}
\newcommand{\LMP}{\mathrm{WMP}}

\newcommand{\Eval}{\textit{Eval}}
\newcommand{\Gval}{\textit{GVAL}}
\newcommand{\Lval}{\textit{WVAL}}
\newcommand{\gval}{\textit{gval}}
\newcommand{\lval}{\textit{wval}}

\newcommand{\Mem}{M}
\newcommand{\prob}{\mathbb{P}}
\newcommand{\Prob}{\textit{Prob}}

\newcommand{\Freq}{\mathit{Freq}}

\newcommand{\Pay}{\textit{Pay}}

\newcommand{\wmp}{\textit{wmp}}

\newcommand{\Dist}{\mathit{Dist}}

\newcommand{\Occ}{\mathit{Occ}}
\newcommand{\gmp}{\mathit{gmp}}
\newcommand{\WinMPsynt}{\textsc{WinMPsynt}}
\newcommand{\ag}[1]{\overline{#1}}

\newcommand{\PTIME}{\mathsf{P}}
\newcommand{\PSPACE}{\mathsf{PSPACE}}
\newcommand{\EXPTIME}{\mathsf{EXPTIME}}
\newcommand{\NP}{\mathsf{NP}}

\newcommand{\qed}{\hfill$\Box$}
\usepackage{xspace}
\makeatletter
\DeclareRobustCommand\onedot{\futurelet\@let@token\@onedot}
\def\@onedot{\ifx\@let@token.\else.\null\fi\xspace}

\def\ie{i.e\onedot}

\makeatother

%% file: sections/abstract.tex
\begin{abstract}
    The long-run average payoff per transition (mean payoff) is the main tool for specifying the performance and dependability properties of discrete systems. The problem of constructing a controller (strategy) simultaneously optimizing several mean payoffs has been deeply studied for stochastic and game-theoretic models. One common issue of the constructed controllers is the \emph{instability} of the mean payoffs, measured by the deviations of the average rewards per transition computed in a finite ``window'' sliding along a run. Unfortunately, the problem of simultaneously optimizing the mean payoffs under local stability constraints is computationally hard, and the existing works do not provide a practically usable algorithm even for non-stochastic models such as two-player games. In this paper, we design and evaluate the first efficient and scalable solution to this problem applicable to Markov decision processes.
\end{abstract}

%% file: sections/intro.tex
\section{Introduction}
\label{sec-intro}

\emph{Mean payoff}, i.e., the long-run average payoff per time unit, is the standard formalism for specifying and evaluating the long-run performance of dynamic systems. The overall performance is typically characterized by a tuple of mean payoffs computed for multiple payoff functions representing expenditures, income, resource consumption, and other relevant aspects. The basic task of \emph{multiple mean-payoff optimization} is to design a controller (strategy) jointly optimizing these mean payoffs. Efficient strategy synthesis algorithms have been designed for various models, such as Markov decision processes or two-player games (see Related work). 

A fundamental problem of the standard mean payoff optimization is the \emph{lack of local stability guarantees}. For example, consider a payoff function modeling the expenditures of a company. Even if the long-run average expenditures per day (mean payoff) are $\$1000$, there are no guarantees on the maximal average expenditures per day in a \emph{bounded time horizon} of, say, one month. It is possible that the company pays between \mbox{$\$800$} and \mbox{$\$1200$} per day every month, which is fine and sustainable. However, it may also happen that there are ``good'' and ``bad'' months with average daily expenditures of $\$0$ and $\$5000$, respectively, where the second case is not so frequent but represents a critical cashflow problem that may ruin the company. The mean payoff does not reflect this difference; hence, optimizing the mean payoff (minimizing the overall long-run expenditures) may lead to disastrous outcomes in certain situations.

The lack of local stability guarantees has motivated the study of \emph{window mean payoff objectives}, where the task is to optimize the average payoff in a ``window'' of finite length sliding along a run. The window size represents a bounded time horizon (in the above example, the horizon of one month), and the task is to construct a strategy such that the average payoff computed for the states in the window stays within given bounds. Thus, one can express both long-run average performance and stability constraints. For a single payoff function, some technical variants of this problem are solvable efficiently, while others are $\PSPACE$-hard. For \emph{multiple} window mean payoffs, intractability results have been established for two-player games (see Related work). 

The main concern of the previous works on window mean payoffs is classifying the computational complexity of constructing an optimal strategy. The obtained algorithms are based on reductions to other problems, and their complexity matches the established lower bounds. To the best of our knowledge, there have been no attempts to tackle the high complexity of strategy synthesis by designing scalable algorithms at the cost of some (unavoidable but acceptable) compromises. This open challenge and the problem's high practical relevance are the primary motivations for our work.

\paragraph{Our Contribution.}

We design the first efficient and scalable strategy synthesis algorithm for optimizing multiple window mean payoffs in a given Markov decision process. 

We start by establishing the principal limits of our efforts by showing that the problem is \mbox{$\NP$-hard} even for simple instances where the underlying MDP is a graph.
Consequently, \emph{every} efficient algorithm attempting to solve the problem must inevitably suffer from some limitations. Our algorithm trades efficiency for optimality guarantees, i.e., the computed solutions are not necessarily optimal. Nevertheless, our experiments show that the algorithm can construct high-quality (and sometimes quite sophisticated) strategies for complicated instances of non-trivial size. Thus, we obtain the first practically usable solution to the problem of multiple window mean payoff optimization.

More concretely, the optimization objective is specified by a function $\Eval$ measuring the ``badness'' of the achieved tuple of window mean payoffs, and the task is to \emph{minimize} the expected value of $\Eval$ (we refer to the next section for precise definitions). The $\Eval$ function can specify complex requirements on the tuple of achieved mean payoffs. This overcomes another limitation of the previous works, where it was only possible to specify the desired upper/lower bounds for each window mean payoff separately.

The core of our strategy synthesis algorithm is a novel procedure based on dynamic programming, computing the distribution of local mean payoffs for a given starting state. Remarkably, this procedure is differentiable, and the corresponding gradient can be calculated (in the backward pass) at essentially the same computational costs. %The details are given in Section~\ref{sec-algo}.

A strategy maximizing the expected value of $\Eval$ may require memory. Our strategy synthesis algorithm produces \emph{finite-memory randomized} strategies, where the memory size is a parameter. Using larger memory may produce better strategies but also substantially increases computational costs. From the scalability perspective, using randomization is \emph{essential} because randomized strategies may achieve much better performance than deterministic strategies with the same amount of memory. This is demonstrated by the simple example below.

\paragraph{Example~1.}
Consider a graph $D$ with two states $A,B$ and two payoff functions where 
$\Pay_1(A) {=} 1$, $\Pay_1(B) {=} 0$, $\Pay_2(A) {=} 0$, and $\Pay_2(B) {=} 8$. We aim to construct a finite-memory strategy such that the pair of expected window mean payoffs in a window of length~$8$ is positive in both components and as close to $(1,1)$ as possible with respect to $L_1$ (Manhattan) distance. Formally, for a given pair of window mean-payoffs $(\wmp_1,\wmp_2)$, the function $\Eval$ returns the $L_1$ distance to the vector $(1,1)$ if both $\wmp_1$ and $\wmp_2$ are positive; otherwise, $\Eval$ returns a ``penalty'' equal to~$5$. 

\begin{figure}[t]\centering
    \begin{tikzpicture}[x=1.5cm, y=2cm, >=stealth', scale=0.48,font=\scriptsize]
        \node  at (-7cm,0cm) {Graph $D$}; 
        \begin{scope}[yshift=0cm]
            \node[min,label={$(1,0)$}] (A) at (0,0) {$A$};
            \node[min,label={$(0,8)$}] (B) at (2,0) {$B$};
            \draw[->,thick] (A) to [in=160,out=20]  (B);
            \draw[->,thick] (B) to [in=340,out=200]  (A);
            \path (A) edge [out=150,in=210,thick, looseness=8, ->] (A);
            \path (B) edge [out=30,in=330,thick, looseness=8, ->] (B);
        \end{scope}
        \node  at (-7cm,-2.5cm) {$K {=} 1$};      
        \begin{scope}[xshift=-4cm, yshift=-2.5cm]
            \node[min] (A) at (0,0) {$A$};
            \node[min] (B) at (2,0) {$B$};
            \draw[->,thick,dashed] (A) to [in=160,out=20]  (B);
            \draw[->,thick,dashed] (B) to [in=340,out=200]  (A);
            \node at (1,-.8) {$\Exp[\Eval] = 3.5$};
        \end{scope}
        \begin{scope}[xshift=4cm, yshift=-2.5cm]
            \node[min] (A) at (0,0) {$A$};
            \node[min] (B) at (2,0) {$B$};
            \draw[->,thick,dashed] (A) to [in=160,out=20] node[above] {$0.27$} (B);
            \draw[->,thick,dashed] (B) to [in=340,out=200]  (A);
            \path (A) edge [dashed,out=150,in=210,thick, looseness=8, ->] node[left] {$0.73$} (A);
            \node at (1,-.8) {$\Exp[\Eval] = 1.43$};
        \end{scope}
        \node  at (-7cm,-7cm) {$K {=} 2$};      
        \begin{scope}[xshift=-4cm, yshift=-7cm]
            \node[min] (A1) at (-.5,0) {$A$};
            \node[min] (A2) at (1,0) {$A$};           
            \node[min] (B) at  (2.5,0) {$B$};
            \draw[->,thick,dashed] (A1) to (A2);
            \draw[->,thick,dashed] (A2) to (B);
            \draw[->,thick,dashed, rounded corners] (B) -- +(0,-.6) -| (A1);
            \node at (1,-1) {$\Exp[\Eval] = 2$};
        \end{scope}
        \begin{scope}[xshift=4cm, yshift=-7cm]
            \node[min] (A1) at (-.5,0) {$A$};
            \node[min] (A2) at (1,0) {$A$};           
            \node[min] (B) at  (2.5,0) {$B$};
            \draw[->,thick,dashed] (A1) to node[above] {$0.45$} (A2);
            \draw[->,thick,dashed] (A2) to node[above] {$0.45$} (B);
            \draw[->,thick,dashed, rounded corners] (B) -- +(0,-.6) -| (A1);
            \node at (1,-1) {$\Exp[\Eval] = 0.94$};
            \path (A1) edge [dashed,out=60,in=120,thick, looseness=8, ->] node[left=.5em] {$0.55$} (A1);
            \path (A2) edge [dashed,out=60,in=120,thick, looseness=8, ->] node[left=.5em] {$0.55$} (A2);
        \end{scope}
        \node  at (-7cm,-11cm) {$K {=} 7$};      
        \begin{scope}[xshift=-5cm, yshift=-11cm, x=1.9cm]
            \foreach \x in {0,1,2,3,4,5,6}{
                \node[min] (A\x) at (\x,0) {$A$};
            }       
            \node[min] (A7) at  (7,0) {$B$};
            \foreach \x/\y in {0/1,1/2,2/3,3/4,4/5,5/6,6/7}{
                \draw[->,thick,dashed] (A\x) to (A\y);
            }
            \draw[->,thick,dashed, rounded corners] (A7) -- +(0,-.6) -| (A0);
            \node at (3.5,-1) {$\Exp[\Eval] = \frac{1}{8}$};
        \end{scope}

    \end{tikzpicture}
    \caption{ 
    For a memory with $K$ states where $K \in \{1,2\}$, the best deterministic strategy (left) is worse than a suitable randomized strategy (right). The optimal finite memory strategy requires $7$ memory elements.}
    \label{fig-graph-mem-helps}
\end{figure}
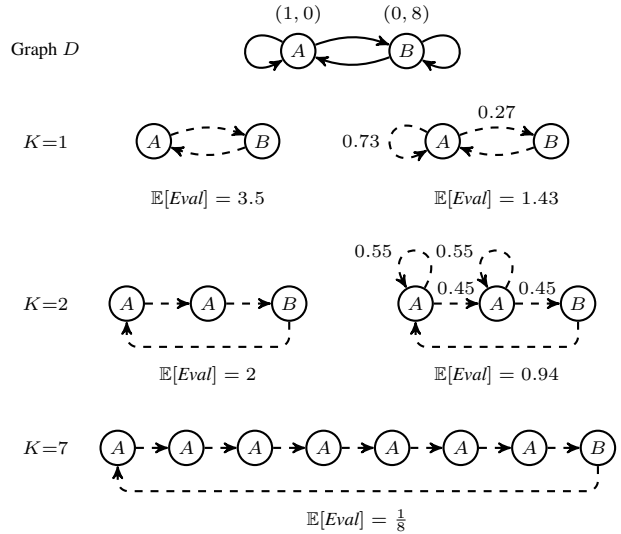

Suppose that the available memory has $K$ different states. Then, for every $K < 7$, there is a \emph{randomized} strategy with $K$ memory states achieving a strictly better expected value of $\Eval$ than the best \emph{deterministic} strategy with $K$ memory states. For $K = 1,2$, such strategies are shown in Fig.~\ref{fig-graph-mem-helps} (for $K = 1$, the best deterministic strategy achieves the window mean payoffs $(\frac{1}{2},4)$ in \emph{every} window of length~$8$, and hence the expected $L_1$ distance to $(1,1)$ is $3.5$; for $K=2$, $2/3$ of the windows have mean payoffs $(\frac{5}{8},3)$, and $1/3$ (those beginning with $AA$) have $(\frac{3}{4},2)$, hence $\Exp[\Eval] = 2$). The \emph{optimal} finite-memory strategy achieving the window mean payoffs $(\frac{7}{8},1)$ in every window of length~$8$ (i.e., the expected value of $\Eval$ equal to~$\frac{1}{8}$) requires $7$~memory elements. 
\qed
\smallskip

In our experiments, we concentrate on the following: 
\begin{itemize}
    \item[(1)]  Demonstrating the improvement achieved by the dynamic procedure described above, where the baseline is a ``naive'' DFS-based procedure.
    \item[(2)] Evaluating the scalability of our algorithm and the quality of the constructed strategies.
    \item[(3)] Analyzing the power of randomization to compensate for insufficient memory.
\end{itemize}
In~(2), a natural baseline for evaluating the quality of a constructed strategy is the best value achievable by a finite-memory strategy. However, there is no algorithm for computing the best value. This issue is overcome by constructing a parameterized family of graphs where the best achievable value and the amount of required memory can be determined by a careful \emph{manual analysis}. 
The structure of our parameterized graphs is similar to the ones constructed in our \mbox{$\NP$-hardness} proofs mentioned above. Thus, we avoid bias towards simple instances of the problem. In~(3), we analyze the quality of randomized strategies constructed for memory of increasing size. Roughly speaking, our experiments show that even in scenarios where the number of memory elements is \emph{insufficient} for implementing an optimal strategy, the quality of the obtained strategies is still relatively close to the optimum. Hence, randomization can ``compensate'' for insufficient memory. This is encouraging because the number of memory elements is one of the crucial parameters influencing the computational costs.

\paragraph{Related Work.} Single mean-payoff optimization for Markov decision processes (MDPs) is a classical problem studied for decades (see, e.g., \cite{Puterman:book}). For multiple mean-payoff optimization in MDPs, a polynomial-time algorithm computing optimal strategies and approximating the Pareto curve has been given in \cite{BBCFK:MDP-two-views}, and the solution has been successfully integrated into the software tool PRISM
\cite{BCFK:Multigain-TACAS}.

The window mean payoff objectives have been first studied for a single payoff function and non-stochastic two-player games in \cite{CHDRR:window-mean-payoff-games-IC}, and then for Markov decision processes in \cite{BGR:window-mean-payoff,BDOR:LifeRandomTimeNot-LMCS}. The complexity of the strategy synthesis algorithms proposed in these works ranges from polynomial time to polynomial space, depending on the concrete technical variant of the problem. The multiple window mean payoff objectives have been considered for two-player games in \cite{CHDRR:window-mean-payoff-games-IC}, where the problem of computing an optimal strategy is classified as provably intractable ($\EXPTIME$-hard). The optimization objective studied in these works is a simple conjunction of upper/lower bounds imposed on the achieved window mean payoffs. The main focus is on classifying the computational complexity of the studied problems, where the upper bounds are typically obtained by reductions to other game-theoretic problems. To the best of our knowledge, our work gives the first scalable algorithm for multiple window mean payoff optimization applicable to Markov decision processes of considerable size.

In a broader context, a related problem of window parity objective has been studied in \cite{CH:finitary-win-omega,Horn:faster-finitary-games,CHH:stoch-fin-obj}.
An alternative approach to capturing local stability of mean payoff based on bounding the variance of relevant random variables has been proposed in \cite{BCFK:performance-stability-JCSS}. 

%% file: sections/model.tex
\section{The Model}
\label{sec-model}

We assume familiarity with basic notions of probability theory (probability distribution, expected value, conditional random variables, etc.) and Markov chain theory. The set of all probability distributions over a finite set~$A$ is denoted by $\Dist(A)$. We use $\Nset$ and $\Qset_{+}$ do denote the set of non-negative integers and non-negative rationals, respectively.

\paragraph{Markov chains.} A \emph{Markov chain} is a triple $M = (S,\Prob,\mu)$ where $S$ is a finite set of \emph{states}, \mbox{$\Prob : S {\times} S \to [0,1]$} is a \emph{stochastic matrix} where $\sum_{t\in S} \Prob(s,t) =1$ for all $s \in S$, and $\mu \in \Dist(S)$ is an \emph{initial distribution}.

A \emph{run} in $M$ is an infinite sequence of states. We use $\prob$ to denote the standard probability measure over the runs of~$M$ (see, e.g., \cite{Norris:book})  

A state $t$ is \emph{reachable} from a state $s$ if $\Prob^n(s,t) > 0$ for some $n \geq 1$.
We say that $M$ is \emph{strongly connected} (or \emph{irreducible}) if all states are mutually reachable from each other. For an irreducible $M$, we use $\Inv \in \Dist(S)$ to denote the unique \emph{invariant distribution} satisfying $\Inv = \Inv \cdot \Prob$ (note that~$\Inv$ is independent of~$\mu$). By ergodic theorem \cite{Norris:book}, $\Inv$ is the limit frequency of visits to the states of~$S$ along a run. More precisely, let $w = s_0,s_1,\ldots$ be a run of~$M$. For every $s \in S$, let 
\[
    \Freq_s(w) =  \lim_{n \to \infty} \frac{\#_s(s_0,\ldots,s_{n-1})}{n}    
\]
where $\#_s(s_0,\ldots,s_{n-1})$ is the number of occurrences of~$s$ in $s_0,\ldots,s_{n-1}$. 
If the above limit does not exist, we put $\Freq_s(w) = 0$. Furthermore, let $\Freq(w) : S \to [0,1]$ be the vector of all $\Freq_s(w)$ where $s \in S$. The ergodic theorem says that \mbox{$\prob[\Freq {=} \Inv] = 1$}. 

A \emph{bottom strongly connected component (BSCC)} of $M$ is a maximal $B \subseteq S$ such that $B$ is strongly connected and closed under reachable states. Note that if $M$ is irreducible, then the set $S$ is the only BSCC of~$M$. Otherwise, $M$ can have multiple BSCCs, and each such $B$ can be seen as an irreducible Markov chain where the set of states is $B$ and the probability matrix is the restriction of $\Prob$ to  $B {\times} B$. 

For the rest of this section, we fix an irreducible Markov chain $M = (S,\Prob,\mu)$.

\paragraph{Global and Local Mean Payoff.}
Let $\Pay : S \to \Nset$ be a \emph{payoff function}. For every run $w = s_0,s_1,\ldots$, let 
\[
    \GMP(w) =  \lim_{n\to \infty} \frac1n \sum_{i=0}^{n-1}\Pay(s_i) 
\]
be the limit-average payoff per transition along the run~$w$. If the limit does not exist, we put $\GMP(w) = 0$. An immediate consequence of the ergodic theorem (see above) is that the defining limit of $\GMP(w)$ exists and takes the same value $\gmp = \sum_{s \in S} \Inv(s) \cdot \Pay(s)$ for almost all runs, i.e., $\prob[\GMP {=} \gmp] = 1$. We refer to the (unique) value $\gmp$ of 
$\GMP$ as the \emph{global mean payoff}.

Let $d \in \Nset$ be a \emph{time horizon}. For every $j \in \Nset$, let
\[
    \LMP[d,j](w) = \frac1d\sum_{i=j}^{j{+}d{-}1} \Pay(s_i)
\]
be the average payoff computed for the sequence of $d$~consecutive states in $w$ starting with~$s_j$. We refer to the value $\LMP[d,j]$ as the \emph{window mean payoff after $j$~steps} (assuming some fixed~$d$).
 
Note that as $d \to \infty$, the value of  $\LMP[d,j]$ is arbitrarily close to $\gmp$ with arbitrarily large probability, independently of~$j$. However, for a given~$d$, the value of $\LMP[d,j]$ depends on~$j$ and can be rather different from~$\gmp$. Also observe that $\LMP[d,j]$ is a discrete random variable, and the underlying distribution depends only on the state~$s_j$. More precisely, for all $j,m \in \Nset$ and $s \in S$ such that $\prob[s_j {=} s] > 0$ and 
$\prob[s_m {=} s] > 0$, we have that the conditional random variables $\LMP[d,j] \,{\mid}\, s_j{=}s$ and $\LMP[d,m] \,{\mid}\, s_{m}{=}s$ are identically distributed. In the following, we write just $\LMP[d,s]$ instead of $\LMP[d,j] \,{\mid}\, s_j{=}s$ where $\prob[s_j {=} s] > 0$.

\paragraph{Mean Payoff Objectives.}

Let $\Pay_1,\ldots,\Pay_k~:~S~\to~\Nset$ be payoff functions. The standard multi-objective optimization problem for Markov Decision Processes (see below) is to jointly maximize the global mean payoffs for $\Pay_1,\ldots,\Pay_k$. In this paper, we use a more general approach where the ``badness'' of the achieved mean payoffs is measured by a dedicated function $\Eval : \Rset^k \to \Rset$. A smaller value of $\Eval$ indicates more appropriate payoffs.

For example, the joint maximization of the mean payoffs can be encoded by $\Eval_{\max}(\vec{x}) = \sum_{i=1}^k (\Pay_i^{\max} {-} \vec{x}_i)$, where $\Pay_i^{\max} = \max_{s \in S} \Pay_i(s)$.
The defining sum  
can also be weighted to reflect certain priorities among the Pareto optimal solutions. 
In general, $\Eval$ can encode the preference of keeping the mean payoffs close to some values, within some interval, or even enforce some mutual dependencies among the mean payoffs. 

As we shall see, our strategy synthesis algorithm works for an arbitrary $\Eval$ that is \emph{decomposable}. Intuitively, the decomposability condition enables more efficient evaluation/differentiation of $\Eval$ by dynamic programming without substantially restricting the 
expressive power.

Now we can define the \emph{global} and \emph{local} value of a run as follows:  
{\small
\begin{align*}
    \Gval(w) & = \Eval(\GMP_1(w),\ldots,\GMP_k(w))\\
    \Lval(w) & = \lim_{n \to \infty} \frac1n \sum_{j=0}^{n-1}\Eval\bigl(\LMP_1[d,j],\ldots,\LMP_k[d,j]\bigr)
\end{align*}}%
where $\GMP_i(w)$ and $\LMP_i[d,j]$ are the global and window mean payoff determined by $\Pay_i$.
Note that $\Lval(w)$ corresponds to the ``limit-average badness'' of the window mean-payoffs along the run~$w$. 

Clearly, $\prob[\Gval {=} \gval] = 1$, where $\gval = \Eval(\gmp_1,\ldots,\gmp_k)$. Furthermore, by applying the ergodic theorem and the above observations leading to the definition of 
$\LMP_i[d,s]$, we obtain that $\prob[\Lval {=} \lval] = 1$ where
{\small%
\begin{equation}
   \lval = \sum_{s \in S} \Inv(s) \cdot \Exp\big[\Eval(\LMP_1[d,s],\ldots,\LMP_k[d,s])\big]
   \label{eq:lval}
\end{equation}}%
We refer to $\gval$ and $\lval$ as the \emph{global} and \emph{window mean payoff value}.
In this paper, we study the problem of minimizing $\lval$ in a given Markov decision process.

\paragraph{Markov decision processes.} A \emph{Markov decision process (MDP)}\footnote{Our definition of MDPs is standard in the area of graph games. Although it is equivalent to the ``classical'' MDP definition of \cite{Puterman:book}, it is more convenient for our purposes and leads to simpler notation.} is a triple 
$D = (V,E,p)$ where $V$ is a finite set of \emph{vertices} partitioned into subsets $(V_N,V_S)$ of \emph{non-deterministic} and \emph{stochastic} vertices, $E \subseteq V {\times} V$ is a set of \emph{edges} s.t.{} each vertex has at least one out-going edge, and $p\colon V_S {\to} \Dist(V)$ is a \emph{probability assignment} such that $p(v)(v') {>} 0$ only if $(v,v') \in E$. We say that $D$ is a \emph{graph} if $V_S {=} \emptyset$.

Outgoing edges in non-deterministic vertices are selected by a \emph{strategy}. In this paper, we consider \emph{finite-memory randomized (FR)} strategies where the selection depends not only on the vertex currently visited but also on some finite information about the history of vertices visited previously.

\paragraph{FR strategies.}
Let $D = (V,E,p)$ be an MDP and $\Mem {\neq} \emptyset$ a finite set of \emph{memory states} that are used to ``remember'' some information about the history. For a given pair $(v,m)$ where $v$ is a currently visited vertex and $m$ a current memory state, a strategy randomly selects a new pair $(v',m')$ such that $(v,v') \in E$. 

Formally, let $\alpha\colon V \to 2^{\Mem}$ be a \emph{memory allocation}, and let 
$\ag{V} = \{(v,m) \mid v \in V, m \in \alpha(v)\}$ be the set of \emph{augmented vertices}. A \emph{finite-memory (FR) strategy} is a function
$\sigma\colon \ag{V} \to \Dist(\ag{V})$ such that for all $(v,m) \in \ag{V}$ where $v \in V_S$ and every $(v,v')\in E$ we have that 
\[
\sum_{m' \in \alpha(v')}\sigma(v,m)(v',m') \ = \ p(v)(v')\,.
\]
An FR strategy is \emph{memoryless} (or \emph{Markovian}) if $\Mem$ is a singleton.  In the following,
we use $\ag{v}$ to denote an augmented vertex of the form $(v,m)$ for some $m\in\alpha(v)$. 

Every FR strategy $\sigma$ together with a probability distribution $\mu \in \Dist(\ag{V})$ determine the Markov chain $D^{\sigma} = (\ag{V},\Prob,\mu)$ where $\Prob(\ag{v},\ag{u}) = \sigma(\ag{v})(\ag{u})$. 

%% file: sections/problem.tex
\section{The Optimization Problem}
\label{sec-problem}

In this section, we define the multiple window mean-payoff optimization problem and examine the principal limits of its computational tractability.

Let $D = (V,E,p)$ be an MDP, $\Pay_1,\ldots,\Pay_k:V{\to}\Nset$ payoff functions, and $\Eval : \Rset^k \to \Rset$ an evaluation function. Furthermore, let $d \in \Nset$ be a time horizon. The task is to construct an FR strategy $\sigma$ and an initial augmented vertex so that the value of $\lval$ is \emph{minimized}. 

More precisely, for a given FR strategy $\sigma$, the function $\lval$ is evaluated as follows.  Every $\Pay_i$ is extended to the augmented vertices of $D^\sigma$ by defining $\Pay_i(\ag{v}) = \Pay_i(v)$. Let $C_1,\ldots,C_n$ be the BSCCs of the Markov chain $D^\sigma$. Recall that every $C_i$ can be seen as an irreducible Markov chain. We use $\lval(C_i)$ to denote the window mean payoff value computed for $C_i$. The value of $\lval$ achieved by $\sigma$, denoted by $\lval^\sigma$, is defined as 
\[
    \lval^\sigma \ = \ \min \{ \lval(C_i)  \mid 1 \leq i \leq n\}  
\]
Recall that the initial augmented vertex can be chosen freely, which is reflected in the above definition (the strategy $\sigma$ can be initiated directly in the ``best'' BSCC and thus achieve the value $\lval^\sigma$).  

A FR-strategy $\sigma$ is \emph{$\varepsilon$-optimal} for a given $\varepsilon \geq 0$ if 
\[
     \lval^\sigma - \varepsilon \ \leq \  \inf_{\pi} \lval^\pi    
\]
where $\pi$ ranges over all FR strategies. A $0$-optimal strategy is called \emph{optimal}. 

The next theorem shows that computing an optimal strategy is computationally hard, even for restricted subsets of instances.

\begin{theorem}
\label{thm-hardness}
    Let $D = (V,E,p)$ be a graph, $\Pay_1,\ldots,\Pay_k$ payoff functions, $d \leq |V|$ a time horizon, and $\Eval$ an evaluation function. The problem of whether there exists a FR strategy $\sigma$ such that $\lval^\sigma \leq 0$ is $\NP$-hard.
    
    Furthermore, the problem is $\NP$-hard even if the set of eligible instances is restricted so that an optimal \emph{memoryless} strategy is guaranteed to exist, and one of the following three conditions is satisfied:
    \begin{itemize}
    \item[A.] $k{=}1$ (i.e., there is only \emph{one} payoff function).
    \item[B.] $k{=}2$, and there are thresholds $c_1,c_2 \geq 0$ such that $\Eval(\kappa_1,\kappa_2) = 0$ iff $\kappa_1 \geq c_1$ and $\kappa_2 \geq c_2$.
    \item[C.] There is a constant $r \in \Nset$ such that $k \leq |V|^{\frac{1}{r}}$ (i.e., the number of payoff functions is ``substantially smaller'' than the number of vertices), and the co-domain of every $\Pay_i$ is $\{0,1\}$.
    \end{itemize}
\end{theorem}   
Intuitively, (A)~says that one payoff function is sufficient for $\NP$-hardness, (B)~says that for two payoff functions we have $\NP$-hardness even if we just aim to push both window mean payoffs above certain thresholds, and (C)~says that the range of all payoff functions can be restricted to $\{0,1\}$ even if the number of payoff functions is at most $|V|^{\frac{1}{r}}$. Furthermore, the $\Eval$ function can be constructed so that it ranges over $\{-t,0\}$ where $t \in \Nset$ is an arbitrary constant, and $\lval^\sigma = 0$ iff $\sigma$ is optimal. Consequently, an \mbox{optimal} (and even \mbox{$t{-}1$-optimal}) strategy cannot be constructed in polynomial time unless $\PTIME = \NP$.

%% file: sections/algorithm.tex
\section{The Algorithm}
\label{sec-algo}

For the rest of this section, we fix an MDP $D = (V,E,p)$, a time horizon $d$, payoff functions $\Pay_1,\ldots,\Pay_k$, and an evaluation function $\Eval$.

Our algorithm is based on optimizing $\lval$ by the methods of differentiable programming.
The core ingredient is a dynamic procedure for computing the expected value of $\Eval(\LMP_1[d,s],\ldots,\LMP_k[d,s])$ (see~\eqref{eq:lval}). The procedure is
designed so that the gradient of $\lval$ can be computed using automatic differentiation. Then, we show how to incorporate this procedure into a strategy-improvement algorithm for $\lval$. 
For the rest of this section, we fix a memory allocation \mbox{$\alpha: V \to 2^M$}.

\paragraph{Representing FR Strategies.} 
\label{sec-sigma-repre}
For every pair of augmented vertices $(\ag{v},\ag{u})$ such that \mbox{$(v,u) \in E$}, we fix a real-valued parameter representing $\sigma(\ag{v})(\ag{u})$. Note that if $\ag{v}$ is stochastic, then the parameter actually represents the probability of selecting the memory state of $\ag{u}$. These parameters are initialized to random values, and we use the standard \textsc{Softmax} function to transform these parameters into probability distributions. Thus, every function $F$ depending on $\sigma$ becomes a function of the parameters, and we use $\nabla F$ to denote the corresponding gradient.

\paragraph{Computing $\lval^\sigma$.}
\label{sec-comp-lval}

Let $\sigma$ be an FR strategy where $\alpha$ is the memory allocation. We show how to compute  $\lval^\sigma$ interpreted as a function of the parameters representing $\sigma$. 

Recall that $\lval^\sigma  = \min \{ \lval(C_i)  \mid 1 \leq i \leq n\}$. Hence, the first step is to compute all BSCCs of $D^\sigma$ by the Tarjan's algorithm \cite{Tarjan:SCC-decomp-SICOMP}. Then, for each BSCC $C$, we compute 
$\lval(C)$ in the following way. 

The invariant distribution $\Inv_C$ is computed as the unique solution of the following system of linear equations: For every $\ag{v}\in C$, we fix a fresh variable $x_{\ag{v}}$ and the corresponding equation $x_{\ag{v}}=\sum_{\ag{u} \in C} x_{\ag{u}}\cdot\sigma(\ag{u})(\ag{v})$. Furthermore, we add the equation
$\sum_{\ag{u} \in C} x_{\ag{u}} = 1$. Recall that $\Inv_C$ is the unique distribution satisfying $\Inv_C = \Inv_C \cdot \Prob_C$, where $\Prob_C$ is the probability matrix of~$C$. Hence, $\Inv_C$ is the unique solution of the constructed system.  

The main challenge is to compute the expected value
\begin{equation}
    \Exp\big[\Eval(\LMP_1[d,\ag{v}^*],\ldots,\LMP_k[d,\ag{v}^*])\big]
\label{eq-Eval-expectation}
\end{equation}
for each $\ag{v}^*\in C$. This is achieved by Algorithm~\ref{alg:localeval} described below. Then, $\lval(C)$ is calculated using~(\ref{eq:lval}). 

\paragraph{Computing the Expected Value of $\Eval$ by Dynamic Programming.}
\label{sec-dynamic}

In this section, we show how to compute~\eqref{eq-Eval-expectation} by dynamic programming for a given $\ag{v}^* \in C$. We use $\prob$ to denote the probability measure over the runs in $C$, where the initial distribution assigns~$1$ to $\ag{v}^*$.

Let $\Nset^{C}$ be the set of vectors of non-negative integers indexed by the augmented vertices $\ag{v}\in C$.
For each $t\in \Nset$, let $\Nset^{C}_t=\{ x\in \Nset^C \mid \ell_x = t\}$, where $\ell_x = \sum_{\ag{u}\in C} x(\ag{u})$.
For every $x \in \Nset^{C}$, let $\Occ_x$ be an indicator assigning to every run 
$w=\ag{v}_0,\ag{v}_1,\ldots$ of $C$ either $1$ or $0$ so that  $\Occ_x(w) = 1$ iff $\#_{\ag{v}} (\ag{v}_0,\ldots \ag{v}_{\ell_x-1}) = x(\ag{v})$ for every $\ag{v} \in C$.

%$\#_\ag{v} (\ag{v}_0,\ldots \ag{v}_{t-1}) = x(\ag{v})$ for all $\ag{v} \in C$ or not, respectively. Here, $\#_\ag{v} (\ag{v}_0,\ldots \ag{v}_{t-1})$ is the number of occurrences of  $\ag{v}$

For every $x \in \Nset^{C}$, let 
\[
   \LMP_i(x) = \frac{\sum_{\ag{v} \in C} x(\ag{v})\cdot\Pay_i(\ag{v})}{\ell_x}.
\] 
Then, \eqref{eq-Eval-expectation} is equal to
\begin{equation}
    \sum_{x \in \Nset^{C}_d} \prob[\Occ_x{=}1] \cdot \Eval(\LMP_1(x),\ldots,\LMP_k(x)).
\label{eq-vectors}
\end{equation}

Calculating~\eqref{eq-vectors} directly is time-consuming. However, $\Eval(\LMP_1(x),\ldots,\LMP_k(x))$ is the same for many different $x \in \Nset^{C}_d$, and our dynamic algorithm avoids these redundant computations. The algorithm is applicable to a subclass of \emph{decomposable} $\Eval$ functions defined below. The decomposability condition is not too restrictive, and it does not influence the \mbox{$\NP$-hardness} of the considered optimization problem (the $\Eval$ functions constructed in the proof of Theorem~\ref{thm-hardness} are decomposable). 

For all $x,y\in \Nset^{C}$ and $\ag{v}\in C$, we write $x\to^{\ag{v}} y$ if $y(\ag{v})=x(\ag{v})+1$ and $y(\ag{u})=x(\ag{u})$ for all $\ag{u} \neq \ag{v}$. We say that $\Eval$ is \emph{decomposable} if there is a set $R$ of \emph{representatives} and efficiently computable functions $r\colon \Nset^{C}\to R$, $e\colon R\to \Rset$ and $m\colon R\times C\to R$ satisfying the following conditions:
\begin{itemize}
    %Pokud zbude dost mista, lze odkomentovat (ale ma to underfull hbox)
    %\item  $r$ assigns a representative to every $x \in \Nset^{C}$ so that $r(x) = r(x')$ implies both $\Eval(\LMP_1(x), \ldots, \LMP_k(x)) = \Eval(\LMP_1(x'), \ldots, \LMP_k(x'))$ and $r(y) = r(y')$ whenever $x\to^{\ag{v}} y$ and $x'\to^{\ag{v}} y'$.
    \item  $\Eval(\LMP_1(x),\dots,\LMP_k(x)) = e(r(x))$ for every $x \in \Nset^{C}_d$, i.e.,
    the value of $\Eval$ for a given $x \in \Nset^C_d$ is efficiently computable by the function $e$ just from the representative of~$x$.
    \item For each $x\to^{\ag{v}} y$, we have that $r(y)=m(r(x),\ag{v})$. That is, when a path is prolonged by a vertex $\ag{v}$, the representative can be efficiently updated by the function~$m$.
\end{itemize}
A concrete example of a decomposable $\Eval$ and the associated $r,e,m$ functions is given in a special subsection below.

For each $t\in \{1,\dots,d\}$, let $R_t = \{r(x) \mid x \in \Nset^C_t \}$. Furthermore, for every representative $\varrho \in R_t$, let $\prob[\varrho] = \sum_{x \in \Nset^C_t\cap r^{-1}(\varrho)}\prob[\Occ_x {=} 1]$.
%and $\LMP_i(\varrho) = \LMP_i(x)$ where $r(x) = \varrho$. %nezda se mi
Then~\eqref{eq-vectors} can be rewritten into
\begin{equation}
    \sum_{\varrho \in R_d} \prob[\varrho] \cdot e(\varrho).
\end{equation}

\begin{algorithm}[t]
	\small
    \caption{Evaluation via dynamic programming}
	\label{alg:localeval}
	\begin{algorithmic}
            \State $\varrho^*$ = $m(r(\{0,\dots,0\}),\ag{v}^*)$
		\State $map_0[(\ag{v}^*,\varrho^*)]$ = $1.$
		\For{$t\in\{1,\dots,d-1\}$}
                \For{$((\ag{v},\varrho),p)\in map_0$}
                    \For{$\ag{u}\in C$}
                        \State $\varrho'$ = $m(\varrho,\ag{u})$
                        \State $map_1[(\ag{u},\varrho')]$ += $p\cdot\sigma[\ag{v}][\ag{u}]$
                    \EndFor
                \EndFor
                \State $swap(map_0,map_1)$
                \State $map_1.clear()$
		\EndFor
        \State $rsl$ = $0.$
        \For{$((\ag{v},\varrho),p)\in map_0$}
            \State $rsl$ += $p\cdot e(\varrho)$
		\EndFor
	\end{algorithmic}
\end{algorithm}

Algorithm~\ref{alg:localeval} computes $\prob[\varrho]$ for all $t\in \{1,\dots,d\}$ and $\varrho\in R_t$ by dynamic programming. % w.r.t.\ $t$. 
Moreover, only reachable representatives (\ie, those with $\prob[\varrho]>0$) are considered during the computation. Thus, Algorithm~\ref{alg:localeval} avoids the redundancies of the direct computation of~\eqref{eq-vectors}.

More specifically, Algorithm~\ref{alg:localeval} uses two associative arrays (such as C++ \textit{unordered\_map}), called $map_0$ and $map_1$, to gather information about the representatives and the corresponding probabilities. In the $t$-th iteration of the cycle, $map_0$ contains items corresponding to all reachable $\varrho \in R_t$, and the items corresponding to all reachable $\varrho \in R_{t+1}$ are gradually gathered in $map_1$. In particular, each $map_i$ is indexed by the elements $(\ag{v},\varrho)\in C\times R$. Intuitively, each such element represents the set of all runs $w = \ag{v}_0,\ag{v}_1,\ldots$ initiated in $\ag{v}^*$ such that there exists $x \in \Nset^C_{t+i}$ satisfying $r(x) = \varrho$, $\Occ_x(w) = 1$, and $\ag{v}_{\ell_x -1} = \ag{v}$. The value associated to $(\ag{v},\varrho)$ is the total probability of all such runs~$w$.

\paragraph{A Simple DFS Procedure.}
\label{sec-naive}

As a baseline for measuring the improvement achieved by the dynamic algorithm described in the previous paragraph, we use a simple \mbox{DFS-based} procedure of Algorithm~\ref{alg:DFS}.
For simplicity, the vector $(\Pay_1(\ag{v}),\dots,\Pay_k(\ag{v}))$ is denoted by $\Pay(\ag{v})$.

The DFS procedure inputs the following parameters:
\begin{itemize}
	\item the current augmented vertex $\ag{v}$;
	\item the probability $p$ of the current path;
	\item the length $n$ of the current path;
	\item the vector $\mathit{payoff\_vector}$ of the individual payoffs accumulated along the path.
\end{itemize}
The procedure is called as DFS($\ag{v}^*,1,1,\Pay(\ag{v})$)
for each $\ag{v}^*$ in the currently examined BSCC $C$.
At the end of the computation, the global variable $\mathit{rsl}$ holds the value of~\eqref{eq-Eval-expectation}.
\begin{algorithm}[h]
	\small
	\caption{Evaluation via DFS}
	\label{alg:DFS}
	\begin{algorithmic}[1]
		\If{$n<d$}
		\For{$\ag{u}\in C$}
		\State DFS($\ag{u},p\cdot\sigma[\ag{v}][\ag{u}],n+1,\mathit{payoff\_vector}+\Pay(\ag{v})$)
		\EndFor
        \Else
        \State $\mathit{rsl}$ += $p\cdot \Eval(\mathit{payoff\_vector})$
		\EndIf
	\end{algorithmic}
\end{algorithm}

\paragraph{A Strategy Improvement Algorithm.}
\label{sec-algorithm}

In this section, we describe a strategy improvement algorithm \mbox{\WinMPsynt} that inputs an MDP $D = (V,E,p)$, payoff functions $\Pay_1,\ldots,\Pay_k : V \to \Nset$, a decomposable $\Eval : \Rset^k \to \Rset$, and a time horizon $d \in \Nset$, and computes an FR strategy $\sigma$ with the aim of minimizing $\lval^\sigma$.

The memory allocation function is a hyperparameter (by default, all memory states are assigned to every vertex). The algorithm proceeds by randomly choosing the parameters representing a strategy. The values are sampled from \mbox{\textsc{LogUniform}} distribution so that no prior knowledge about the solution is imposed. 
Then, the algorithm computes the BSCCs of $D^\sigma$ and identifies a BSCC~$C$ with the best $\lval(C)$. Subsequently, $\lval(C)$ is improved by gradient descent. The crucial ingredient of \WinMPsynt\ is Algorithm \ref{alg:localeval},
allowing to compute $\lval(C)$ and its gradient by automatic differentiation. After that,  the point representing the current strategy is updated in the direction of the steepest descent. 
The intermediate solutions and the corresponding $\lval(C)$ values are stored, and the best solution found within \textsc{Steps} optimization steps is returned (the value of \textsc{Steps} is a hyper-parameter). Our implementation uses \textsc{PyTorch} framework \cite{PyTorch} and its automatic differentiation with \textsc{ADAM} optimizer \cite{adam}.
Observe that \WinMPsynt\ is equally efficient for general MDPs and graphs. The only difference is that stochastic vertices generate fewer parameters.

\paragraph{An Example of a Decomposable \Eval\ Function.}
\label{sec-decomp}
%In this section, we illustrate the notion of a decomposable evaluation function on a simple example.  

Let $\Pay$ be a payoff function, and let $\Eval : \Rset \to \Rset$ where $\Eval(P)$ is either $0$ or $1$ depending on whether $P \in [3,5]$ or not, respectively. Assume $d = 5$. Then, we can put $R = \{0,\dots,26\}$ and define
\begin{eqnarray*}
   r(x) & = & \min\{a,26\}, \mbox{ where } a= \sum_{\ag{v} \in C} x(\ag{v}) \cdot \Pay(\ag{v})\\[-1ex]
   e(\varrho) & = & \begin{cases}
                        0 & \mbox{if } \varrho/d \in [3,5],\\
                        1 & \mbox{otherwise.}
                    \end{cases}\\
    m(\varrho,\ag{v}) & = & \min\{\varrho{+}\Pay(\ag{v}), 26\}
\end{eqnarray*}   
Note that as soon as the accumulated payoff exceeds $25$, there is no reason to remember the exact value because $\Eval$ inevitably becomes one. Note that $R$ contains $27$ elements \emph{independently} of the size of $C$, while the total number of all $x \in \Nset^C_5$ such that $\prob[\Occ_x {=} 1] > 0$ may exceed $\binom{|C|}{4}$, and the total number of paths, all of which are considered separately by the naive DFS-based algorithm, may reach $|C|^4$.

%% file: sections/experiments.tex
\section{Experiments}
\label{sec-experiments}

We perform our experiments on graphs to separate the probabilistic choice introduced by the constructed strategies from the internal probabilistic choice performed in stochastic vertices. The graphs are structurally similar to the ones constructed in the \mbox{$\NP$-hardness} proof of Theorem~\ref{thm-hardness}. This avoids bias towards simple instances. Recall that the problem of constructing a (sub)optimal strategy for such graphs is \mbox{$\NP$-hard} even if just one memory state is allocated to every vertex, there are only two payoff functions, and we aim at pushing the window mean payoffs above certain thresholds (see item~B.{} in Theorem~\ref{thm-hardness}).

\begin{figure}\centering
\tikzstyle{st}=[thick,circle,draw,minimum size=1.6em,inner sep=0em,text centered]
\tikzstyle{ar}=[thick,stealth-] 
\begin{tikzpicture}[font=\scriptsize,scale=0.65]
\foreach \i/\l in {0/0, 1/60, 2/120, 3/180, 4/240, 5/300}{
    \node[st] (n1\i) at (\l:1.5) {$\textcolor{red}{10},\!\textcolor{blue}{0}$};
    \node[st] (n2\i) at (\l:2.7) {$\textcolor{red}{2},\!\textcolor{blue}{2}$};
    \node[st] (n3\i) at (\l:3.9) {$\textcolor{red}{0},\!\textcolor{blue}{10}$};
}
\foreach \i/\j in {0/1, 1/2, 2/3, 3/4, 4/5, 5/0}{
    \draw[ar] (n1\i) -- (n1\j); 
    \draw[ar] (n2\i) -- (n1\j);
    \draw[ar] (n1\i) -- (n2\j); 
    \draw[ar] (n2\i) -- (n2\j);
    \draw[ar] (n3\i) -- (n2\j);
    \draw[ar] (n2\i) -- (n3\j); 
    \draw[ar] (n3\i) -- (n3\j);
}
\end{tikzpicture}
\caption{The graph $D_6$ and the payoffs assigned to vertices.
}
\label{fig-ring}
\end{figure}
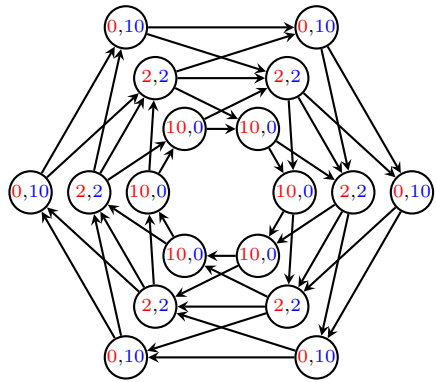

\paragraph{The graphs $\pmb{D_\ell}$.}
For every $\ell \geq 2$, we construct a directed ring $D_\ell$ with three ``layers'' where every vertex in the inner, middle, and outer layer is assigned a pair of payoffs
$(10,0)$, $(2,2)$, and $(0,10)$, respectively. The vertices are connected in the way shown in Fig.~\ref{fig-ring}. 

\paragraph{The $\pmb{\Eval}$ function.}
A \emph{scenario} is a pair $(\ell,d)$ where $\ell,d$ are even integers in the interval $[2,20]$ representing $D_\ell$ and the window length~$d$. For every scenario, we aim to push both window mean payoffs simultaneously above a bound $b$, where $b$ is as large as possible. For each 
$(\ell,d)$, the \emph{maximal} bound achievable by an FR strategy is denoted by $b_{\ell,d}$, and can be determined by a careful \emph{manual analysis}, together with the least number of memory states $K_{\ell,d}$ required by an optimal FR strategy    
(we have that $K_{\ell,d} \leq 5$ for all $(\ell,d)$).  Let us note that the manual analysis of $D_\ell$ is enabled by the regular structure of $D_\ell$, but this regularity does not bring \emph{any advantage} to {\WinMPsynt}.
 
For a given scenario $(\ell,d)$, we use the evaluation function $\Eval_{\ell,d}$ defined as follows:
\[
  \Eval_{\ell,d}(P_1,P_2) = \frac{\max\{0,b_{\ell,d} -P_1\} + \max\{b_{\ell,d} - P_2\}}{2\cdot b_{\ell,d}}.
\] 
By the definition of $b_{\ell,d}$, there always exists an optimal FR strategy $\sigma$ for the scenario $(\ell,d)$ achieving the bound $b_{\ell,d}$, i.e., $\lval^{\sigma} = 0$. Due to the normalizing denominator, the maximal value of $\Eval_{\ell,d}$ is bounded by~$1$, simplifying the comparison of strategies constructed for different scenarios.

\paragraph{The experiments.} 

For all scenarios $(\ell,d)$ and $K \in \{1,\dots,5\}$, we invoked 
{\WinMPsynt} $100$~times, where the underlying memory allocation assigns $K$ memory elements to every vertex, and the number of optimization steps is set to $1000$. Thus, for every $(\ell,d)$ and every $K$, we obtained a set $\Sigma_{\ell,d,K}$ of $100$ strategies and the corresponding values. We also use $\Sigma_{\ell,d}$ to denote $\bigcup_{K=1}^5 \Sigma_{\ell,d,K}$, and $\Sigma$ to denote the union of all $\Sigma_{\ell,d}$.

\paragraph{The quality of the obtained strategies.} 
Due to the definition of $\Eval_{\ell,d}$, for every $\sigma \in \Sigma_{\ell,d}$, the value $\lval^{\sigma}$ can be interpreted as a normalized distance to the \emph{optimal} strategy with value~$0$. The percentage of scenarios where the value of the \emph{best} strategy found by 
\mbox{\WinMPsynt} is bounded by $0$, $0.05$, and $0.1$ is $40\%$, $52\%$, and $100\%$, respectively. 
Hence, the best strategies found by \mbox{\WinMPsynt} are of \emph{very high quality}. Since the best strategy is selected out of $500$ strategies constructed for a given scenario, a natural question is how good are these strategies ``on average''. The percentage of all $\sigma \in \Sigma$ whose value is bounded by $0$, $0.1$, and $0.3$ is $6\%$, $27\%$, and $99\%$, respectively. Hence, the quality of an ``average'' strategy is substantially worse, which is consistent with intuitive expectations (since the problem is computationally hard, obtaining a high-quality solution for non-trivial instances cannot be easy). %More details are in \suppl.

\begin{figure}
  \includegraphics[scale=0.37]{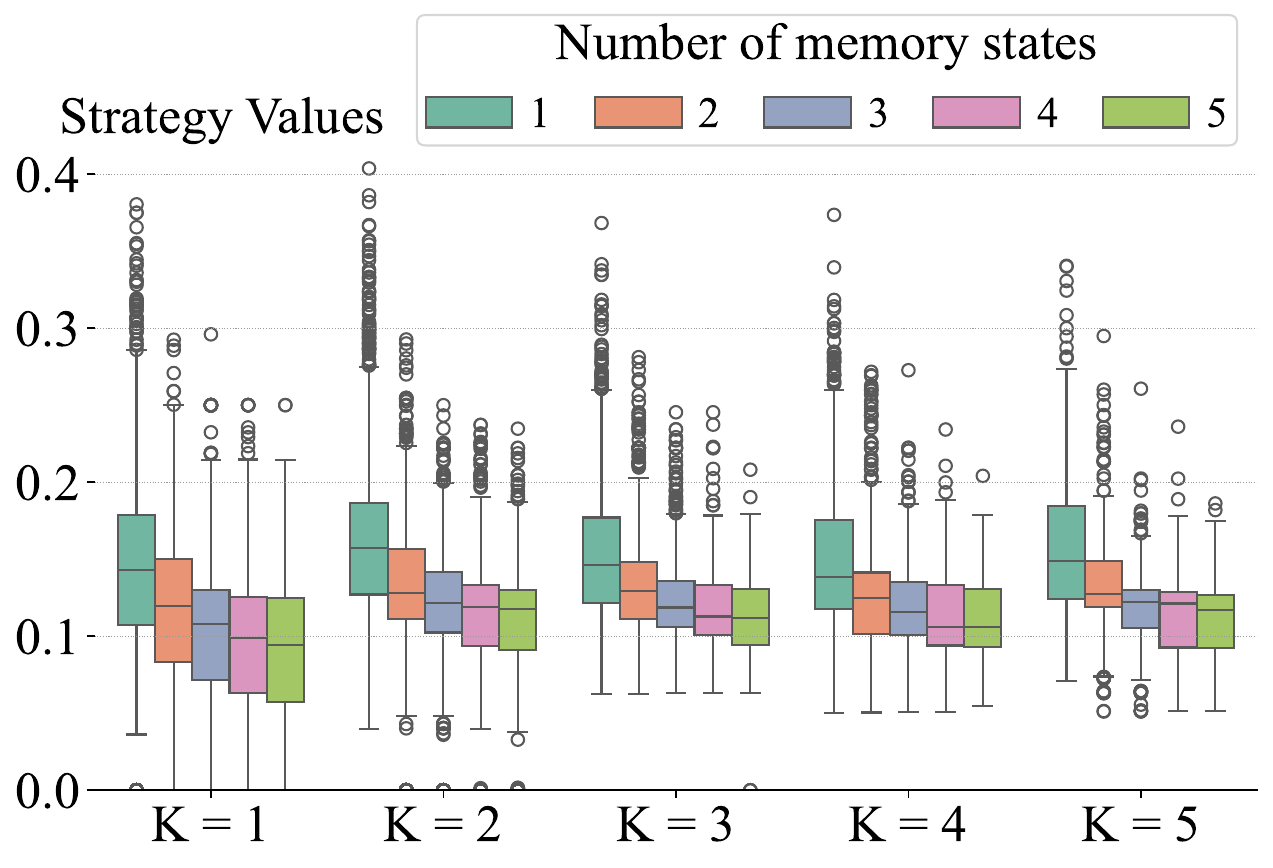}
  \caption{More memory states lead to better strategies.}
  \label{fig-memory}
\end{figure}

The roles of memory states and randomization are demonstrated in Fig.~\ref{fig-memory}.
The scenarios are split into five disjoint subsets with the same $K_{\ell,d}$ (horizontal axis). For each subset, we report the values achieved by strategies with $1,\ldots,5$ memory states assigned to every vertex (indicated by different colors). Note that for the subset where $K_{\ell,d} = 2$, an optimal strategy is computed when $2$ or more memory states are available. For the subset where $K_{\ell,d} = 3$, an optimal strategy is found only for $5$ memory states. 
For all subsets, increasing the number of memory states decreases the average strategy value. 
Furthermore, even if the number of memory states is smaller than $K_{\ell,d}$, the value of the constructed strategies is still relatively small on average. Hence, 
randomization effectively compensates for the lack of memory.

\paragraph{The improvement achieved by dynamic programming.} The baseline for evaluating the efficiency improvement achieved by the dynamic procedure of Algorithm~\ref{alg:localeval} is the simple DFS-based procedure of Algorithm~\ref{alg:DFS}. For every instance $(\ell,d)$ where $4 \leq \ell {=} d \leq 30$, we report the average running time of one training step for an FR strategy with $1,\ldots,5$ memory states (different colors) using \emph{logarithmic} scale. The timeout is set to $20$ secs. For one memory state, the DFS-based procedure reaches the timeout for all scenarios $(\ell,d)$ where $\ell {=} d \leq 20$, whereas the dynamic procedure needs less than one second even for the $(30,30)$ scenario. Hence, the dynamic procedure 
substantially outperforms the  DFS-based one, and the same holds when the number of memory states increases.

\begin{figure}
  \includegraphics[scale=0.37]{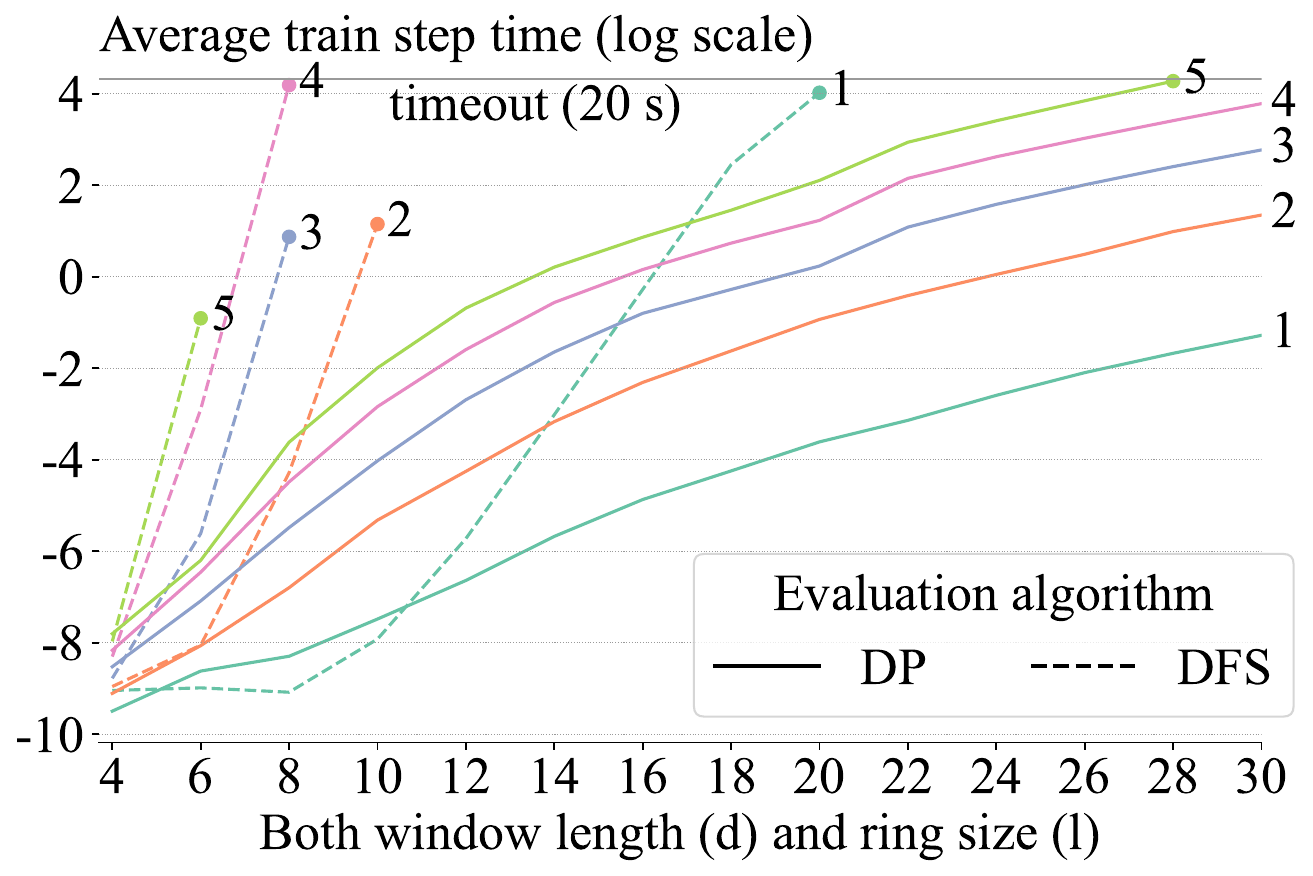}
  \caption{Dynamic evaluation procedure outperforms the DFS-based one.}
  \label{fig-times}
\end{figure}

%% file: sections/conclusion.tex
\section{Conclusions}
We have designed an efficient strategy synthesis algorithm for optimizing multiple window mean payoffs capable of producing high-quality solutions for instances of considerable size.
An interesting question is whether the proposed approach is applicable to a larger class of window-based optimization objectives such as the window parity objectives.

%% file: sections/appendix.tex
\section{A Proof of Theorem~1}

Let us start by restating the theorem.

\begin{theorem}
    Let $D = (V,E,p)$ be a graph, $\Pay_1,\ldots,\Pay_k$ payoff functions, $d \leq |V|$ a time horizon, and $\Eval$ an evaluation function. The problem of whether there exists a FR strategy $\sigma$ such that $\lval^\sigma \leq 0$ is $\NP$-hard.
        
    Furthermore, the problem is $\NP$-hard even if the set of eligible instances is restricted so that an optimal \emph{memoryless} strategy is guaranteed to exist, and one of the following three conditions is satisfied:
    \begin{itemize}
        \item[A.] $k{=}1$ (i.e., there is only \emph{one} payoff function).
        \item[B.] $k{=}2$, and there are thresholds $c_1,c_2 \geq 0$ such that $\Eval(\kappa_1,\kappa_2) = 0$ iff $\kappa_1 \geq c_1$ and $\kappa_2 \geq c_2$.
        \item[C.] There is a constant $r \in \Nset$ such that $k \leq |V|^{\frac{1}{r}}$ (i.e., the number of payoff functions is ``substantially smaller'' than the number of vertices), and the co-domain of every $\Pay_i$ is $\{0,1\}$.
    \end{itemize}
\end{theorem}   

\noindent
For every $\ell \geq 1$, let $D_\ell = (V_\ell,E_\ell,p_\ell)$ be the graph where 
\begin{itemize}
    \item $V_{\ell} = \{s_i,t_i,u_i \mid 0 \leq i < \ell\}$;
    \item $E_\ell$ contains the edges $(s_i,t_i)$, $(s_i,u_i)$,  $(t_i,s_{i+})$, $(u_i,s_{i+})$
      for every $0 \leq i < \ell$, where $s_{i+} = s_{(i+1) \bmod \ell}$.
    \item since $D_\ell$ is a graph, the probability assignment $p_\ell$ is empty.
\end{itemize}

\begin{figure}[tbh]
\centering
    \tikzstyle{st}=[thick,circle,draw,minimum size=1.6em,inner sep=0em,text centered]
    \tikzstyle{ar}=[thick,-stealth] 
    \begin{tikzpicture}[font=\scriptsize,scale=0.9]
    \foreach \i/\s/\u in {0/180/150, 1/120/90, 2/60/30, 3/0/330, 4/300/270, 5/240/210}{
        \node[st] (s\i) at (\s:2.5) {$s_{\i}$};
        \node[st] (t\i) at (\u:1.7) {$t_{\i}$};
        \node[st] (u\i) at (\u:2.8) {$u_{\i}$};
    }
    \foreach \i/\j in {0/1, 1/2, 2/3, 3/4, 4/5, 5/0}{
        \draw[ar] (s\i) -- (u\i); 
        \draw[ar] (s\i) -- (t\i); 
        \draw[ar] (t\i) -- (s\j); 
        \draw[ar] (u\i) -- (s\j); 
    }
    \end{tikzpicture}
    \caption{The graph $D_6$.
    % The values of $\Pay_1$ and $\Pay_2$ are in red and blue.
    }
\label{fig-tworing}
\end{figure}
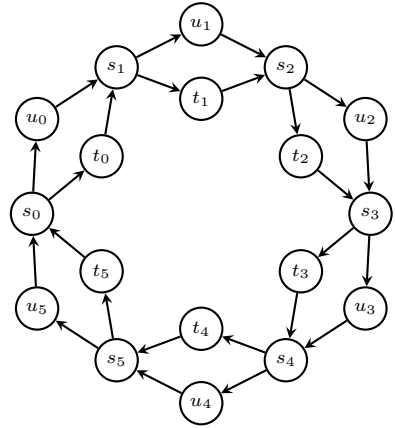

The structure of $D_6$ is shown in Fig.~\ref{fig-tworing}. The cases~A.--C.{} are considered separately.

\paragraph{Case A.} An instance of the \textsc{Subset Sum} problem is a sequence of positive integers $n_0,\ldots,n_{\ell-1}$ and another positive integer $T$. The question is whether this sequence contains a subsequence such that the sum of all elements in the subsequence is equal to~$T$. The \textsc{Subset Sum} is one of the standard \mbox{$\NP$-complete problems} (all $n_j$ are written in binary)

For a given instance $n_0,\ldots,n_{\ell-1}$, $T$ of \textsc{Subset Sum}, consider the graph $D_\ell$ and the payoff function $\Pay$ such that 
\begin{itemize}
    \item $\Pay(s_0) = - T$, and $\Pay(s_i) = 0$ for all $1 \leq i < \ell$,
    \item $\Pay(u_i) = n_i$ and $\Pay(t_i) = 0$ for all $0 \leq i < \ell$.
\end{itemize}
Observe that $n_0,\ldots,n_{\ell-1}$ contains a subsequence with a sum of $T$ iff there is a path of the form $s_0,y_0,s_1,y_1,s_2,\ldots y_{\ell-1}$ where every $y_i$ is either $t_i$ or $u_i$ such that the total payoff accumulated along this path is equal to~$0$. 

We show that the latter condition is equivalent to the existence of a FR strategy $\sigma$ such that $\lval^\sigma \leq 0$, where $d = 2\ell$ and $\Eval(\kappa)$ is the absolute value of $\kappa$. This proves the \mbox{$\NP$-hardness} of Case~A.

Clearly, if there is a path $s_0,y_0,s_1,y_1,s_2,\ldots y_{\ell-1}$ such that every $y_i$ is either $t_i$ or $u_i$ and the total payoff accumulated along this path is equal to~$0$, then there exists a \emph{memoryless} strategy $\sigma$ for $D_\ell$ such that $\lval^{\sigma} = 0$ ($\sigma$ simply follows the path forever, using the edge $(y_{\ell-1},s_0)$).

Now let $\sigma$ be an FR strategy such that $\lval^\sigma \leq 0$. Then, $\lval^\sigma = 0$ because $\Eval$ is non-negative. Let $C$ be a BSCC of $D_{\ell}^{\sigma}$ such that $\lval(C) = 0$. Due to the structure of $D_\ell$, there is an augmented vertex of the form $\ag{s_0} \in C$. Then, $\ag{s_0}$ together with the next $2\ell -1$ augmented vertices visited with some positive probability inevitably form a path of the form $\ag{s_0},\ag{y_0},\ag{s_1},\ag{y_1},\ag{s_2},\ldots \ag{y_{\ell-1}}$ where every $y_i$ is either $t_i$ or $u_i$. Observe that the total payoff accumulated along this path must be zero (if it was positive or negative, then $\lval(C) > 0$, and we have a contradiction). Since the underlying path $s_0,y_0,s_1,y_1,s_2,\ldots,y_{\ell-1}$ has the same accumulated payoff equal to~$0$, we are done.

\paragraph{Case B.} Here we use a reduction from the \mbox{$\NP$-complete} \textsc{Knapsack} problem. An instance of \textsc{Knapsack} is a finite sequence $(v_0,w_0),\ldots,(v_{\ell-1},w_{\ell-1})$, where all $v_i$ and $w_i$ are positive integers, together with two positive integer bounds $V,W$. The question is whether there exists a subsequence such that the sum of all $v_i$'s is at least $V$, and the sum of all $w_i$'s is at most~$W$.
We assume that $\sum_{i=0}^{\ell-1}w_i>W$ (otherwise the instance could be solved trivially by checking whether $\sum_{i=0}^{\ell-1}v_i\geq V$).

For a given instance $(v_0,w_0),\ldots,(v_{\ell-1},w_{\ell-1})$, $V$, $W$ of \textsc{Knapsack}, let $w_{\max}=\max \{w_i\mid 0\leq i<\ell\}$, and consider the graph $D_\ell$ and two payoff functions $\Pay_1,\Pay_2$ such that

\begin{itemize}
    \item $\Pay_1(s_i) = 0$, $\Pay_2(s_i) = w_{\max}$ for all $0 \leq i < \ell$,
    \item $\Pay_1(u_i) = v_i$, $\Pay_2(u_i) = w_{\max} - w_i$ for all $0 \leq i < \ell$,
    \item $\Pay_1(t_i) = 0$, $\Pay_2(t_i) = w_{\max}$ for all $0 \leq i < \ell$.
\end{itemize}

Let $d = 2\ell$, $c_1=V/d$, $c_2=w_{\max}-W/d$, and
\[
  \Eval(\kappa_1,\kappa_2) = \begin{cases}
                                0 & \mbox{if $\kappa_1 \geq c_1$ and $\kappa_2 \geq c_2$}\\
                                1 & \mbox{otherwise.}
                             \end{cases}
\]
Clearly, $c_1>0$, and $d\cdot w_{\max}>\ell\cdot w_{\max}\geq \sum_{i=0}^{\ell-1}w_i>W$, hence $c_2>0$.

Similarly as in Case~A., one can easily verify that 
\begin{itemize}
    \item if $(v_0,w_0),\ldots,(v_{\ell-1},w_{\ell-1})$, $V$, $W$ is a positive instance of \textsc{Knapsack}, then there exists a \emph{memoryless} startegy $\sigma$ such that $\lval^\sigma = 0$.
    \item if there exists a FR strategy $\sigma$ such that $\lval^\sigma = 0$, then $(v_0,w_0),\ldots,(v_{\ell-1},w_{\ell-1})$, $V$, $W$ is a positive instance of \textsc{Knapsack}.
\end{itemize}

\paragraph{Case C.}  The \textsc{Sat} problem is the question of whether a given propositional formula is satisfiable. The $\textsc{Sat}$ problem is \mbox{$\NP$-complete} even for formulae of the form $C_1 \wedge \cdots \wedge C_k$ where each $C_j$ is a disjunction of literals over $x_0,\ldots,x_{\ell-1}$ (a literal is a propositional variable or its negation). Furthermore, the problem remains \mbox{$\NP$-complete} if we additionally assume that $k \leq \ell^{\frac{1}{r}}$, where $r \in \Nset$ is an arbitrary constant.

For a given formula $C_1 \wedge \cdots \wedge C_n$ over the propositional variables $x_0,\ldots,x_{\ell-1}$, consider the graph $D_\ell$ and the payoff functions $\Pay_1,\ldots,\Pay_k$ such that, for every $1\leq j \leq k$, the following holds:
\begin{itemize}
    \item $\Pay_j(s_i) = 0$ for all $0 \leq i < \ell$;
    \item for all $0 \leq i < \ell$, we have that $\Pay_j(u_i)$ equals $1$ or $0$, depending on whether $x_i$ occurs in the clause $C_j$ or not, respectively;
    \item for all $0 \leq i < \ell$, we have that $\Pay_j(t_i)$ equals $1$ or $0$, depending on whether $\neg x_i$ occurs in the clause $C_j$ or not, respectively.
\end{itemize}
Observe that the co-domain of $\Pay_j$ is $\{0,1\}$, and $k \leq |V|^{\frac{1}{r}}$ because $|V| \geq \ell$. 

Let $\Eval(\kappa_1,\ldots,\kappa_k)$ be either $0$ or $1$ depending on whether each $\kappa_j$ is positive or not, respectively. Furthermore, let $d = 2\ell$. Now it is easy to check the following:
\begin{itemize}
    \item If $C_1 \wedge \cdots \wedge C_n$ is satisfiable, then there exists a path in $D_\ell$ of the form $s_0,y_0,s_1,y_1,\ldots,y_{\ell-1}$ such that the vector of accumulated total payoffs is positive in every component.
    Hence, there is a memoryless strategy $\sigma$ such that $\lval^\sigma = 0$.
    \item If there is FR strategy $\sigma$ such that $\lval^\sigma \leq 0$, then there is a  path in $D_\ell$ of the form $s_0,y_0,s_1,y_1,\ldots,y_{\ell-1}$ such that the vector of accumulated total payoffs is positive in every component (here we argue similarly as in Case~A.) Note that this path determines a satisfying assignment $\nu$ for the formula, where $\nu(x_i)$ is true iff the vertex $u_i$ occurs in the path.
\end{itemize}
\bigskip

\noindent
Note that in all of the considered cases, the $\Eval$ function can be trivially adjusted so that it returns either $0$ or a given constant $t \in \Nset$, depending on whether the given payoff (or tuple of payoffs) corresponds to a satisfying subsequence/assignment or not, respectively. This implies that the problem of constructing a $t-1$-optimal strategy is also computationally hard.

\section{Experiments}

In this section, we give more details about the experiments. The quality of the best strategies constructed by \mbox{\WinMPsynt} for each scenario is reported in Table~\ref{tab-best}. More precisely, the table shows the percentage of scenarios where the value of the best strategy $\sigma^*$ found by \mbox{\WinMPsynt} is below a given threshold.
\begin{table}
\centering
\begin{tabular}{ll}
    \toprule
    Bound on $\lval^{\sigma^*}$ & $\%$ of scenarios\\
    \midrule
    0.0  & 40\,\% \\ 
    0.01 & 42\,\% \\
    0.02 & 42\,\% \\
    0.03 & 43\,\% \\
    0.04 & 49\,\% \\
    0.05 & 52\,\% \\
    0.06 & 66\,\% \\
    0.07 & 78\,\% \\
    0.08 & 90\,\% \\
    0.09 & 98\,\% \\
    0.1 & 100\,\% \\
     \bottomrule
\end{tabular}
\caption{The percentage of scenarios where the value $\lval^{\sigma^*}$ of the best strategy $\sigma^*$ found by \mbox{\WinMPsynt} for a given scenario is less than or equal to a given bound.}
\label{tab-best}
\end{table}

Similar statistics for the set $\Sigma$ of \emph{all} of the $50.000$ strategies 
constructed by \mbox{\WinMPsynt} for all scenarios are given in Table~\ref{tab-average}. 

As we already noted, the best strategies constructed by \mbox{\WinMPsynt} are of very high quality. The value of an ``average'' strategy is not so close to the optimum due to the high computational complexity of the considered optimization problem.

\begin{table}
    \centering
    \begin{tabular}{ll}
        \toprule
        Bound on the value & $\%$ of strategies\\
        \midrule
        0.0 & 6\,\% \\  
        0.1 & 27\,\% \\ 
        0.2 & 95\,\% \\ 
        0.3 & 99.8\,\% \\ 
        0.4 & 99.9\,\% \\ 
        0.5 & 100\,\% \\
        \bottomrule
    \end{tabular}
\caption{The percentage of strategies constructed by \mbox{\WinMPsynt} with the value less than or equal to a given bound.}
\label{tab-average}
\end{table}

\begin{theorem}
    Let $\ell, d$ be even positive integers. Then $b_{\ell, d} = \frac{10(d/2 - 1) + 4}{d}$.
\end{theorem}
We will first show that $b_{l, d} \geq \frac{10(d/2 - 1) + 4}{d}$. Let $k = \frac{d}{2} - 1$, since $d$ is even and positive, then $k$ is a non-negative integer. Consider a FR strategy that starts in the middle layer. It then moves along the inner layer $k$ times, moves back into the middle layer, and moves $k$ times in the outer layer, finally it returns back to the middle layer. This pattern has a total length equal to $k + 1 + k + 1 = d$. If we repeat this pattern $\mathrm{lcm}(l, d) / d$ times, we get a cycle of length $\mathrm{lcm}(l, d)$ which can be easily represented as an FR strategy.

Moreover, in each window of length $d$ we have exactly $10 k + 4$ on both payoffs. Hence we can satisfy a bound $b = \frac{10 k + 4}{d}$.

The other inequality is easy to see, as one would need to sacrifice going in the inner layer for going more in the outer or vice versa.

\paragraph{Values of $K_{\ell, d}$}
The analysis for $K_{\ell, d}$ was done by an algorithm that computes the memory needed for the optimal strategy described above.
%The code can be found in the file: \url{synthesis/modules/window_mean_payoff/graph.py}, see class ExperimentIWMPOptimalMemory.

The values for $K_{\ell, d}$ when both $\ell$ and $d$ are even positive integers $\leq 20$ are reported in Table~\ref{tab:max_mem}.
\begin{table}[tbh]
    \centering
\begin{tabular}{lrrrrrrrrrr}
\toprule
d = & 2 & 4 & 6 & 8 & 10 & 12 & 14 & 16 & 18 & 20 \\
\midrule
r=20 & 1 & 1 & 1 & 2 & 1 & 2 & 3 & 2 & 4 & 1 \\
r=18 & 1 & 2 & 1 & 2 & 2 & 2 & 3 & 4 & 1 & 5 \\
r=16 & 1 & 1 & 1 & 1 & 2 & 2 & 3 & 1 & 4 & 3 \\
r=14 & 1 & 2 & 1 & 2 & 2 & 3 & 1 & 4 & 4 & 5 \\
r=12 & 1 & 1 & 1 & 2 & 2 & 1 & 3 & 2 & 2 & 3 \\
r=10 & 1 & 2 & 1 & 2 & 1 & 3 & 3 & 4 & 4 & 2 \\
r=8 & 1 & 1 & 1 & 1 & 2 & 2 & 3 & 2 & 4 & 3 \\
r=6 & 1 & 2 & 1 & 2 & 2 & 2 & 3 & 4 & 2 & 5 \\
r=4 & 1 & 1 & 1 & 2 & 2 & 2 & 3 & 2 & 4 & 3 \\
r=2 & 1 & 2 & 1 & 2 & 2 & 3 & 3 & 4 & 4 & 5 \\
\bottomrule
\end{tabular}
    \caption{Maximal memory needed for the optimal FR strategy.}
    \label{tab:max_mem}
\end{table}